# ARTICULATORY INFORMATION AND MULTIVIEW FEATURES FOR LARGE VOCABULARY CONTINUOUS SPEECH RECOGNITION

*Vikramjit Mitra\*, Wen Wang, Chris Bartels, Horacio Franco, Dimitra Vergyri*

University of Maryland, College Park, MD, USA
Speech Technology and Research Laboratory, SRI International, Menlo Park, CA, USA
vmitra@umd.edu, {wen.wang, chris.bartels, horacio.franco, dimitra.vergyri}@sri.com

## ABSTRACT

This paper explores the use of multi-view features and their discriminative transforms in a convolutional deep neural network (CNN) architecture for a continuous large vocabulary speech recognition task. Mel-filterbank energies and perceptually motivated forced damped oscillator coefficient (DOC) features are used after feature-space maximum-likelihood linear regression (fMLLR) transforms, which are combined and fed as a multi-view feature to a single CNN acoustic model. Use of multi-view feature representation demonstrated significant reduction in word error rates (WERs) compared to the use of individual features by themselves. In addition, when articulatory information was used as an additional input to a fused deep neural network (DNN) and CNN acoustic model, it was found to demonstrate further reduction in WER for the Switchboard subset and the CallHome subset (containing partly non-native accented speech) of the NIST 2000 conversational telephone speech test set, reducing the error rate by 12% relative to the baseline in both cases. This work shows that multi-view features in association with articulatory information can improve speech recognition robustness to spontaneous and non-native speech.

***Index Terms***— multi-view features, feature combination, large vocabulary continuous speech recognition, robust speech recognition, articulatory features

## 1. INTRODUCTION

Spontaneous speech typically contains a significant amount of variation, which makes it difficult to model in automatic speech recognition (ASR) systems. Such variability stems from varying speakers, pronunciation variations, speaker stylistic differences, varying recording conditions and many other factors. Recognizing words from conversational telephone speech (CTS) can be quite difficult due to the spontaneous nature of speech, its informality, speaker variations, hesitations, disfluencies etc. The Switchboard and Fisher [1] data collections are large collection of CTS datasets that have been used extensively by researchers working on conversational speech recognition [2, 3, 4, 5, 6]. Recent trends in speech recognition [7, 8, 9] have demonstrated impressive performance on Switchboard and Fisher data.

Deep neural network (DNN) based acoustic modeling has become the state-of-the-art in automatic speech recognition (ASR) systems [10, 11]. It has demonstrated impressive performance gains for almost all tried languages and acoustic conditions. Advanced variants of DNNs, such as convolutional neural nets (CNNs) [12], recurrent neural nets (RNNs) [13], long short-term memory nets (LSTMs) [14], time-delay neural nets (TDNNs) [15, 29], VGG-nets [8], have significantly improved recognition performance, bringing them closer to human performance [9]. Both abundance of data and sophistication of deep learning algorithms have symbiotically contributed to the advancement of speech recognition performance. The role of acoustic features has not been explored in comparable detail, and their potential contribution to performance gains is unknown. This paper focuses on acoustic features and investigates how their selection improves recognition performance using benchmark training datasets: Switchboard and Fisher, when evaluated on the NIST 2000 CTS test set [2].

We investigated a traditional CNN model and explored the following:
(1) Use of multiple features both in isolation and in combination.
(2) Explored different ways of using the feature space maximum-likelihood regression (fMLLR) transform, where we tried (a) learning the fMLLR transforms directly using the filterbank features and (b) learning the fMLLR transform on the cepstral version of the features and then performing inverse discrete cosine transform (IDCT) on the fMLLR features to generate the fMLLR version of filterbank features.
(3) Investigated the use of articulatory features, where the features represent a time series definition of how the vocal tract shape and constrictions change over time.

Our experiments demonstrated that the use of feature combinations helped to improve performance over individual features in isolation and over traditionally used mel-filterbank (MFB) features. Articulatory features were found to be useful for improving recognition performance on both Switchboard and CallHome subsets of the NIST 2000 CTS test set. These findings indicate that the use of better acoustic features can help improve speech recognition performance when using standard acoustic modeling techniques, and can demonstrate performance as good as those obtained from more sophisticated acoustic models that exploit temporal memory. For the sake of simplicity, we used a CNN acoustic model in our experiment, where the baseline system's performance is directly comparable to the state-of-the-art CNN performance reported in [8]. We expect our results using the CNN to carry over into other neural network architectures as well.

---
*\*The author performed this work while at SRI International and is currently working at Apple Inc.*

The outline of the paper is as follows. In Section 2 we present the dataset and the recognition task. In Section 3 we describe the acoustic features and the articulatory features that were used in our experiments. Section 4 presents the acoustic and language models used in our experiments, followed by experimental results in Section 5 and conclusion and future directions in Section 6.

## 2. DATA AND TASK

The acoustic models in our experiments were trained using the CTS Switchboard (SWB) [16] and Fisher (FSH) corpora. We first investigated contributions of the features on models trained only with the SWB dataset, where the training data consisted of ~360 hours of speech data. We then evaluated the contributions of the features using acoustic models trained with a combination of both SWB and FSH (~2000 hours). The models were evaluated using the NIST 2000 CTS test set, which consists of 2.1 hours (21.4K words, 40 speakers) of SWB audio and 1.6 hours (21.6K words, 40 speakers) of the CallHome (CH) audio. The language model training data included 3M words from Switchboard, CallHome, and Switchboard Cellular transcripts, 20M words from Fisher transcripts, 150M words from Hub4 broadcast news transcripts and language model training data, and 191M words of "conversational" text retrieved from the Web by searching for conversational n-grams extracted from the CTS transcripts [25]. A 4-gram language model (LM) was generated based on word probability estimates from a SuperARV language model, which is a class-based language model with classes derived from Constraint Dependency Grammar parses [26]. For first pass decoding the 4-gram LM was pruned to improve efficiency, and the full 4-gram LM was used to rescore lattices generated from the first pass.

## 3. FEATURES

We used mel-filterbank energies (MFBs) as the baseline feature, where the features were generated using the implementation distributed with the Kaldi toolkit [17]. The second acoustic feature was Damped Oscillator Coefficients (DOCs) [18]. The DOC features model the auditory hair cells using a bank of forced damped oscillators, where gammatone filtered band-limited subband speech signals are used as the forcing function. The oscillation energy from the damped oscillators was used as the DOC features after power-law compression.

We performed the fMLLR transform on the acoustic features, where we trained Gaussian Mixture Models (GMMs) to generate alignments on the training dataset to learn the fMLLR transform for the feature sets. We investigated two approaches: (1) we directly learned the fMLLR transforms on the 40-dimensional filterbank features, and (2) we investigated learning the fMLLR transform using the cepstral version of the features. The cepstral version of the features helps decorrelate the features, which in turn adheres to the diagonal covariance assumption of the GMMs. In (2) the fMLLR transform was learned using 40 dimensional cepstral features (using all the cepstral dimensions extracted from 40 dimensional filterbanks). After the fMLLR transform was performed, an IDCT of the features was obtained to generate the fMLLR version of filterbank features.

The articulatory features were estimated using the CNN system described in [19, 20], where the CNN performs speech-to-articulatory inversion or simply speech-inversion. During speech-inversion, the acoustic features extracted from the speech signal, in this case modulation features [19], are used to predict the articulatory trajectories. The articulatory features contain time domain articulatory trajectories, with eight dimensions reflecting: glottal aperture, velic opening, lip aperture, lip protrusion, tongue tip location and degree, tongue body location and degree. More details regarding the articulatory features and their extraction are provided in [19].

## 4. RECOGNITION SYSTEM

We trained CNN acoustic models for the speech recognition tasks. To generate the alignments necessary for training the CNN system, a Gaussian Mixture Model - Hidden Markov Model (GMM-HMM) based acoustic model was first trained with flat-start, which was used to produce the senone labels. Altogether, the GMM-HMM system produced 5.6K context-dependent (CD) states for the SWB training set. A fully connected DNN model was then trained using MFB features, which in turn was used to generate the senone alignments to train the baseline and other acoustic models presented in this work. The input features to the acoustic models were formed using a context window of 15 frames (7 frames on either side of the current frame).

The acoustic models were trained by using cross-entropy (CE) followed by sequence training using maximum mutual information (MMI) criterion [17, 21]. For the CNN model, 200 convolutional filters of size 8 were used in the convolutional layer, and the pooling size was set to 3 without overlap. The subsequent, fully connected network had five hidden layers, with 2048 nodes per hidden layer, and the output layer included as many nodes as the number of CD states for the given dataset. The networks were trained using an initial four iterations with a constant learning rate of 0.008, followed by learning-rate halving based on the cross-validation error decrease. Training stopped when no further significant reduction in cross-validation error was noted or when cross-validation error started to increase. Backpropagation was performed using stochastic gradient descent with a mini-batch of 256 training examples.

In this work, we investigated a modified deep neural network architecture to jointly model the acoustic and the articulatory spaces, as shown in Figure 1. In this modified architecture, two parallel input layers are used to accept acoustic features and articulatory features. The input layer tied to the acoustic feature consists of a convolutional layer, with 200 filters and the input layer tied to the articulatory features is a feed-forward layer with 100 neurons. The feature maps from the convolutional layer and the outputs from the feed-forward layer are fed to a fully connected

DNN consisting of 5 hidden layers and 2048 neurons in each layer, as shown in figure 1.

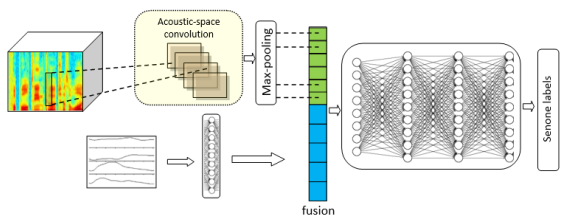

Figure 1. Fused CNN-DNN acoustic model. The convolution input layer accepts acoustic features as input and the feed-forward input layer accepts articulatory features (vocal tract constriction (TV) variables) as input.

## 5. RESULTS

We initially validated the performance of the features (MFB, DOC and TVs) using the 360 hours SWB training dataset. The baseline DNN and CNN models had six and five hidden layers respectively, with 2048 neurons in each layer, and were trained with MFB features and its fMLLR transformed version (MFB+fMLLR). The NIST RT-04 dev04 dataset (3 hour test set from Fisher, containing 36 conversations) [2] was used as the cross-validation set during the acoustic model training step. Table 1 presents the word error rates (WER) from the baseline CNN model trained with the SWB data when evaluated on the NIST 2000 CTS test set, for both cross-entropy (CE) training and sequence training (ST) using MMI. Table 1 also shows the results obtained from the DOC features with and without a fMLLR transform. We present results from ST as they were found to be always better than the results CE training. We explored learning the fMLLR transform directly from the filterbank features (MFB_fMLLR and DOC_fMLLR) and learning the fMLLR transforms on the full dimensional cepstral versions of the features, applying the transform and then performing IDCT (MFB+fMLLR and DOC+fMLLR).

Table 1. WER from the 360 hours SWB trained ST acoustic models when evaluated on the NIST 2000 CTS test set, for MFB and DOC features respectively.

| Feature | Model | WER SWB | WER CH |
|---|---|---|---|
| MFB | DNN | 13.5 | 26.2 |
| DOC | DNN | 12.6 | 23.7 |
| MFB_fMLLR | DNN | 11.8 | 22.2 |
| MFB+fMLLR | DNN | 11.6 | 21.9 |
| DOC_fMLLR | DNN | 12.3 | 23.2 |
| DOC+fMLLR | DNN | 12.0 | 22.9 |
| MFB+fMLLR | CNN | **11.3** | 21.8 |
| DOC+fMLLR | CNN | **11.3** | **20.6** |

Table 1 shows that the performance of fMLLR transforms learned from the cepstral version of the features are better than the ones directly from the filterbank features, which is expected, as the cepstral features are uncorrelated, which adheres to the diagonal covariance assumption of the GMM models used to learn those transforms. Table 1 demonstrates that the fMLLR transformed features always performed better than the features without fMLLR transform. Also, the CNN models always gave better results, confirming similar observations from studies reported earlier [8]. Also, note that Table 1 shows that the DOC features performed slightly better than the MFB features after the fMLLR transform, where the performance improvement was more pronounced for the CH subset of the NIST 2000 CTS test set.

As a next step, we investigated the efficacy of feature combination and focused only on the CNN acoustic models. We appended the articulatory features (TVs) extracted from the SWB training set, dev04 and NIST 2000 CTS test sets, and combined them with MFB+fMLLR and DOC+fMLLR features, respectively. Finally, we combined the MFB+fMLLR and DOC+fMLLR features and added the TVs to them. Table 2 presents the WERs obtained from evaluating all the models trained with different combinations of features. Note that all models using TVs used the fused CNN-DNN (f-CNN-DNN) architecture shown in Figure 1, for jointly modeling the dissimilar acoustic and articulatory spaces. When combining the MFB+fMLLR and DOC+fMLLR features, we trained a CNN model instead. The number of convolutional filters in all the experiments was kept at 200, and only the patch size was increased from eight to twelve in the case of combined acoustic features (MFB+fMLLR + DOC+fMLLR) as opposed to the individual acoustic features (i.e., MFB+fMLLR or DOC+fMLLR).

Table 2. WER from the 360 hours SWB trained ST acoustic model when evaluated with the NIST 2000 CTS test set, for different feature combinations.

| Feature | Model | WER SWB | WER CH |
|---|---|---|---|
| MFB+fMLLR + TV | f-CNN-DNN | 11.2 | 20.8 |
| DOC+fMLLR + TV | f-CNN-DNN | 11.0 | 20.5 |
| MFB+fMLLR + DOC+fMLLR | CNN | 10.7 | 20.4 |
| MFB+fMLLR + DOC+fMLLR +TV | f-CNN-DNN | **10.5** | **19.9** |

Table 2 shows that the use of articulatory features helped to lower the WER in all the cases. The DOC feature was always found to perform slightly better than the MFBs and the best results were obtained when all the features were combined together, indicating the benefit of using multi-view features. Note that only 100 additional neurons were used to accommodate the TV features, hence all the models were of comparable sizes. The benefit of the articulatory features stemmed from the complementary information that they contain (reflecting degree and location of articulatory constrictions in the vocal tract), as demonstrated by earlier studies [22-24]. Overall the f-CNN-DNN system trained with the combined feature set, MFB+fMLLR + DOC+fMLLR + TV, demonstrated a relative reduction in WER of 7% and 9% compared to the MFB+fMLLR CNN baseline for SWB and CH subsets of the NIST 2000 CTS test set. Table 1 and 2 also demonstrates that sequence training always gave additive performance gain over cross-entropy training, supporting the in [8, 21].

As a next step, we focused on training the acoustic models using the 2000-hour SWB+FSH CTS data, focusing on the CNN acoustic models and multi-view features. Note that the MFB DNN baseline model was used to generate the alignments for the FSH part of the 2000 hours CTS training set and as a consequence the number of senone labels remained the same as the 360-hour SWB models. Table 3 presents the results from the 2000 hours CTS trained models. The model configurations and their parameter size were kept the same as the 360-hour SWB models.

Figure 3 shows that the use of the additional FSH training data resulted in significant performance improvement for both SWB and the CH subsets of the NIST 2000 CTS test set. Adding the FSH dataset resulted in relative WER reduction of 4.4% and 12% respectively for SWB and CH subsets of the NIST 2000 CTS test set, using MFB+fMLLR features. Similar improvement was observed from the DOC+fMLLR features as well, where 8% and 12% relative reduction in WER for SWB and CH subsets was observed when FSH data was added to the training data. Note that the CH subset of the NIST 2000 CTS test set was more challenging than the SWB subset, as it contains non-native speakers of English, hence introducing accented speech into the evaluation set. The use of articulatory features helped to reduce the error rates for both SWB and CH test sets, indicating their robustness to model spontaneous speech in both native (SWB) and non-native (CH) speaking styles. The FSH corpus contains speech from quite a diverse set of speakers, helping to reduce the WER of the CH subset more significantly than the SWB subset, a trend reflected in results reported in the literature [8].

Table 3. WER from the 2000 hours SWB+FSH trained acoustic model when evaluated on the NIST 2000 CTS test set, for different feature combinations.

| Feature | Model | WER SWB | WER CH |
|---|---|---|---|
| MFB+fMLLR | CNN | 10.8 | 19.2 |
| DOC+fMLLR | CNN | 10.4 | 18.1 |
| MFB+fMLLR + DOC+fMLLR | CNN | 9.8 | 17.2 |
| MFB+fMLLR + DOC+fMLLR +TV | f-CNN-DNN | **9.5** | **16.9** |

Table 3 demonstrates the benefit of using multi-view features, where a CNN trained with MFB+fMLLR and DOC+fMLLR resulted in reducing the WER by 6% and 5% relatively, for SWB and CH evaluation sets respectively, when compared to the best single feature system DOC+fMLLR. When the articulatory features in the form of the TVs were used in addition to the MFB+fMLLR and DOC+fMLLR features in a f-CNN-DNN model, the best performance from a single acoustic model was obtained, which produced a relative WER reduction of 3% and 2% for SWB and CH evaluation sets respectively, compared to the CNN acoustic model trained with MFB+fMLLR and DOC+fMLLR features.

Table 4 shows the system fusion results after dumping 2000-best lists from the rescored lattices from each individual system of different front-end features with fMLLR, i.e., MFB, DOC, MFB+DOC, MFB+DOC+TV, then conducting M-way combination of the subsystems using N-best ROVER [27] implemented in SRILM [28]. In this system fusion experiment, all subsystems have equal weights for N-best ROVER. As can be seen from the table, N-best ROVER based 2-way and 3-way system fusion produced a further 2% and 4% relative reduction in WER compared to the best single system (MFB+fMLLR + DOC+fMLLR + TV), for SWB and CH evaluation sets respectively. Note that the first row of Table 4 is the last row of Table 3, i.e., the best single system. The last row 4-way fusion is from combining the 4 individual systems presented in Table 3.

Table 4. WER from system fusion experiments.

| System Fusion | WER SWB | WER CH |
|---|---|---|
| Best Single System | 9.5 | 16.9 |
| Best 2-way fusion | **9.3** [MFB+DOC, MFB+DOC+TV] | 16.4 [MFB+DOC, MFB+DOC+TV] |
| Best 3-way fusion | 9.3 [MFB, MFB+DOC, MFB+DOC+TV] | **16.3** [MFB, DOC, MFB+DOC+TV] |
| 4-way fusion | 9.3 | 16.7 |

## 6. CONCLUSION

We reported the results exploring multiple features for ASR on English CTS data. We observed that the fMLLR transform helped reduce the WER of the baseline system significantly. We observed that using multiple acoustic features helped in improving the overall accuracy of the system. Use of robust features and articulatory features significantly reduced the WER for the more challenging CallHome subset of the NIST 2000 CTS evaluation set, with accented speech in that subset. We developed a fused-CNN-DNN architecture, where input convolution was only performed on the acoustic features and the articulatory features were process by a feed-forward layer. We found this architecture effective for combining acoustic features and articulatory features. The robust features and articulatory features capture complementary information, and the addition of them resulted in the best single system performance, with 12% relative reduction of WER on SWB and CH evaluation sets respectively, compared to the MFB+fMLLR CNN baseline.

Note that in this study the language model has not been optimized. Future studies should investigate RNN or other neural network-based language modeling techniques that are known to perform better than word n-gram LMs. Also, advanced acoustic modeling, through the use of time-delayed neural nets (TDNNs), long short-term memory neural nets (LSTMs), and the VGG nets, should also be explored as their performance has been mostly reported using MFB features, and the use of multi-view features can help further improve their performance.